\documentclass[sigconf, review=false]{acmart}

\usepackage{booktabs} 

\usepackage{tabularx}
\usepackage{booktabs}
\usepackage{xcolor}
\usepackage{hyperref}
\usepackage{pgfplotstable}
\usepackage{caption}
\setcopyright{rightsretained}

\copyrightyear{2024}
\acmYear{2024}
\setcopyright{rightsretained}
\acmConference[ICVGIP 2024]{Indian Conference on Computer Vision Graphics and Image Processing}{December 13--15, 2024}{Bengaluru, India}
\acmBooktitle{Indian Conference on Computer Vision Graphics and Image Processing (ICVGIP 2024), December 13--15, 2024, Bengaluru, India}
\acmPrice{}
\acmDOI{10.1145/3702250.3702292}
\acmISBN{979-8-4007-1075-9/24/12}

\begin{document}
\title[Refine3DNet]{Refine3DNet: Scaling Precision in 3D Object Reconstruction from Multi-View RGB Images using Attention}
\titlenote{Produces the permission block, and
  copyright information}

\author{Ajith Balakrishnan}
\affiliation{
  \institution{Center for Interdisciplinary Research}
  \city{College of Engineering Trivandrum}
  \state{Kerala}
  \country{India}
}
\email{tve17ecra02@cet.ac.in}
\email{ajithkannampara@gmail.com}

\author{Dr Sreeja S}
\affiliation{%
  \institution{ Department of Electrical and Electronics Engineering}
  \city{College of Engineering Trivandrum}
  \state{Kerala}
  \country{India}
}
\email{sreejas@cet.ac.in}

\author{Dr. Linu Shine}
\affiliation{
  \institution{Department of Electronics \& Communication Engineering}
  \city{College of Engineering Trivandrum}
  \state{Kerala}
  \country{India}
}
\email{linushine@cet.ac.in}


\renewcommand{\shortauthors}{}

\begin{abstract}
Generating 3D models from multi-view 2D RGB images has gained significant attention, extending the capabilities of technologies like Virtual Reality, Robotic Vision, and human-machine interaction. In this paper, we introduce a hybrid strategy combining CNNs and transformers, featuring a visual auto-encoder with self-attention mechanisms and a 3D refiner network, trained using a novel Joint Train Separate Optimization (JTSO) algorithm. Encoded features from unordered inputs are transformed into an enhanced feature map by the self-attention layer, decoded into an initial 3D volume, and further refined. Our network generates 3D voxels from single or multiple 2D images from arbitrary viewpoints. Performance evaluations using the ShapeNet datasets show that our approach, combined with JTSO, outperforms state-of-the-art techniques in single and multi-view 3D reconstruction, achieving the highest mean intersection over union (IOU) scores, surpassing other models by 4.2\% in single-view reconstruction.
\end{abstract}

%
%



\maketitle

\section{Introduction}
The field of research is becoming more and more interested in reconstructing 3D scenes from 2D images. This is because of the recent availability of large catalogs of 3D models that enable new possibilities for 3D reasoning from photographs. Reconstruction using 3D scans automatically enables quick development of 3D objects and significantly expands the capabilities of immersive technologies like Virtual Reality (VR), Ultrasound (US) \cite{b21}, Robotic Vision,
medical research, military applications, Computed Tomography (CT) \cite{b22}, satellite and remote sensing \cite{b23}, Positron Imaging Tomography (PIT) \cite{b24}, Magnetic Resonance Imaging (MRI), and entertainment. In real-world applications, the 2D to 3D conversion techniques that calculate the depth map from 2D scenes for 3D reconstruction offer a viable way to reduce the expense of encoding, sending, and storing 3D visual material. 3D modeling software can be complex and time-consuming for those with little to no prior experience in 3D modeling.

\par

Conventional techniques, such as Structure from Motion (SFM) \cite{b14} and Visual Simultaneous Localization and Mapping (vSLAM) \cite{b13}, match image features across views. These methods fail if a large baseline separates the image viewpoints due to changes in local appearance or occlusions within the scene \cite{b27}. The reflecting behavior of the object surface also makes the output 3D model noisier. 3D voxel reconstruction methods, such as space carving, have become popular because they can reconstruct 3D structures even with a large baseline problem. These techniques rely on the assumption that the cameras are calibrated or that the objects are precisely separated from the backdrop, which is not always the case. Recent approaches are based on CNNs or transformers \cite{b31}, each with its advantages and limitations. CNN architectures are efficient in single-view 3D reconstruction, while transformers excel in multi-view reconstruction, feature detailing, and managing memory footprints effectively. CNN feature fusion with unordered input images is a challenging task \cite{b26}.

\par
\begin{figure*}[ht]
\centering
\includegraphics[width=.9\textwidth , height=5cm]
{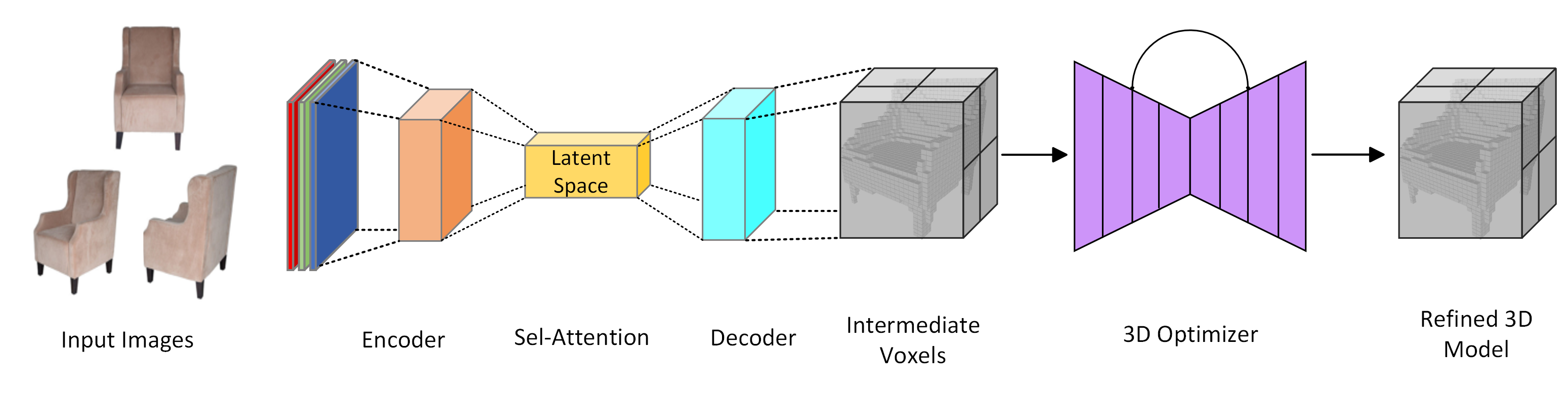}
\caption{ Visualization of Refine3DNet for 3D Object Reconstruction from multiple images.}
\label{fig_fig1}
\end{figure*}

The recent methods, such as 3D-R2N2 \cite{b2}, LSM \cite{b10}, Deep MVS \cite{b12}, and Pix2Vox \cite{b4}, estimate 3D shapes from multiple images with promising results. These methods utilize a shared encoder-decoder mechanism, known as an auto-encoder. 3D-R2N2 \cite{b2} and LSM \cite{b10} employ RNNs to fuse multiple 2D image features extracted by the shared encoder. However, RNN-based methods have three main drawbacks: permutation variance, long-term memory loss \cite{b25}, and lack of parallelization \cite{b29}, which hinder reliable 3D geometry estimation, full exploitation of input images, and increase inference time, respectively. Deep-MVS \cite{b12} uses max pooling to gather deep features but only captures first-order information, leading to a loss of significant image details. \citeauthor{b4}\cite{b4} proposed a shared encoder-decoder architecture followed by an optimizer network based on 3D U-Net \cite{b28} and a VGG16-based encoder-decoder architecture. Nonetheless, both networks fail to process small image features.

\par


This work proposes a novel network architecture for 3D model reconstruction and a dedicated training algorithm to improve training efficiency. The proposed model architecture is based on a shared CNN encoder-decoder architecture followed by a 3D U-net for feature refinement and a dedicated algorithm named JTSO for training. As shown in Fig-\ref{fig_fig1}, the basic block diagram of the proposed CNN architecture reconstructs 3D models from 2D images. To address the feature fusion problem, an attention network inspired by the Transformer model \cite{b19} is used for transforming latent space features. The network employs a ResNet18-based autoencoder architecture for feature extraction and object generation. Refiner networks based on 3D U-Net \cite{b28} augment and optimize the 3D structures generated by decoders \cite{b15,b4}. This network aims to produce highly accurate and scalable 3D representations. We evaluated and compared the network using several state-of-the-art techniques and datasets, demonstrating its efficacy with the specified training algorithm.

The following is a summary of our approach's primary contributions:
\begin{itemize}

\item We introduce an innovative CNN architecture for 3D model reconstruction capable of generating 3D voxels from one or multiple 2D images captured from various viewpoints.

\item We introduce an innovative self-attention mechanism designed to effectively aggregate features from unordered images.

\item We develop a novel 3-stage training algorithm that decouples the network and efficiently updates parameters in distinct stages.

\item Our comprehensive evaluation using the ShapeNetCore dataset demonstrates that the methods proposed in this study surpass state-of-the-art results, particularly with a small number of input images.

\end{itemize}

This is how the remainder of the paper is structured: A brief history of related work is given in Section 2. Our comprehensive network design is presented in Section 3. The suggested training algorithm and training set are covered in Section 4. At last, we showcase the outcomes of our methodology's tests on the ShapeNet and online product datasets in Section 5.

\par

\section{Related Work}
Previous research in the field of 3D model reconstruction has explored various methodologies aimed at enhancing the accuracy, efficiency, and inference time of generated 3D model representations. These approaches range from single-view reconstruction to more complex multi-view methods. In this section, we delve into the developments in this area, focusing on the different methodologies and the challenges they address.

\subsection{Multi-view 3D Model Reconstruction}

One classical computer-vision challenge is extracting three-dimensional information from images. Conventional techniques include visual Simultaneous Localization and Mapping (vSLAM) \cite{b13} and Structure from Motion (SfM) \cite{b14} to match manually crafted features across several views and estimate the camera pose for each image. When the input photos are separated by a long baseline, matching becomes very difficult, and these approaches only work well under favorable conditions.

\par
In response to the limitations of traditional methods, deep learning approaches have emerged, utilizing extensive datasets like ShapeNet \cite{b6} and Pascal 3D+ \cite{b7}. Based on these datasets, several efficient deep-learning methods for 3D reconstruction have been developed. Most of these techniques utilize a shared encoder-decoder design. GRU units are used in both the 3D-R2N2 \cite{b2} and LSM \cite{b10} techniques to fuse deep image features and create 3D models, which introduces permutation variance and gradient issues. Besides the encoder-decoder structure, LSM \cite{b10} includes a 3D Grid Reasoning module for generating final 3D models and depth maps. Methods like DeepMVS \cite{b12} and 3DensiNet use pooling operations to fuse image features extracted by the shared encoder. However, pooling operations capture only first-order information, resulting in the loss of valuable information.

\par


\begin{figure*}[ht]
\centering
\includegraphics[width=.85\textwidth , height=7.3cm]
{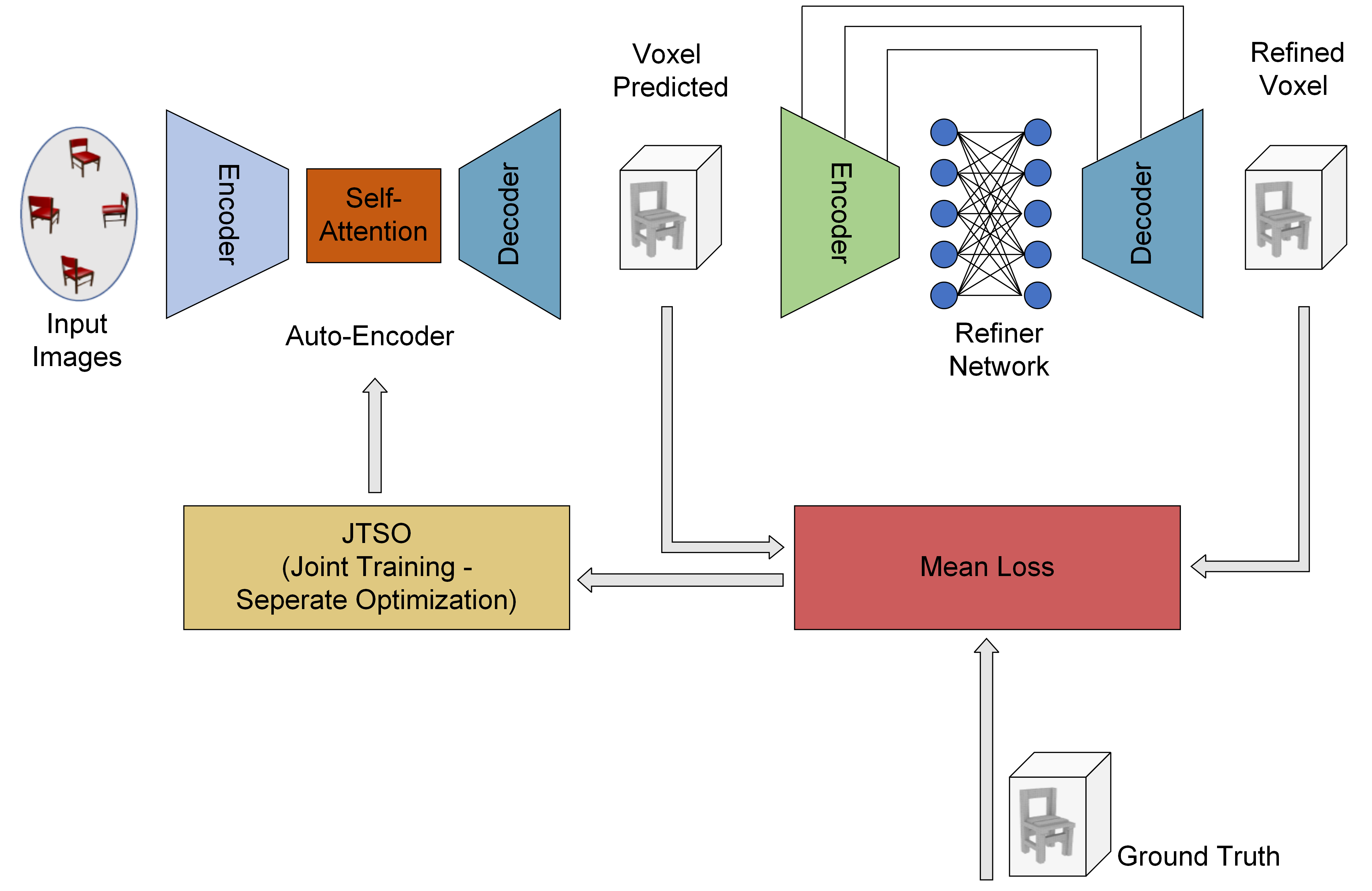}
\caption{Detailed architecture of the proposed network}
\label{fig_fig3}
\end{figure*}

Pix2Vox \cite{b4} utilizes a VGG16-based \cite{b17} shared encoder-decoder architecture for each image, with generated 3D models fused using a context-aware fusion mechanism. Additionally, it includes a Refiner module to optimize the predicted 3D model, which is particularly effective with a low number of input images. The context-aware fusion mechanism, as opposed to RNN modules, allows Pix2Vox to produce more accurate 3D models, although it may struggle with reconstructing small features.


The paper \citeauthor{b1}\cite{b1}  presented a module for aggregating attention to fuse features. 
This module employs a fully connected layer to generate attention scores for each feature, aiding in achieving permutation invariance. The Attsets module replaces the RNN module and pooling operations, leading to improved reconstruction of objects in the form of a 3D occupancy grid. Compared to other networks, Attsets has fewer learnable parameters, minimizing inference time. However, the attention module, termed Attsets-FC, while simpler and less resource-intensive, struggles with complex tasks and is slower due to its sequential processing nature.

\par

The paper \citeauthor{b31}\cite{b31}  proposes a novel framework called 3D Volume Transformer (VolT) for generating 3D models from 2D images. The central approach utilizes an encoder for 2D perspectives and a decoder for 3D volumes. The 2D-view encoder leverages view attention layers to capture information across multiple views, while the 3D-volume decoder utilizes volume attention layers to understand spatial correlations and predict the 3D structure. The framework uniformly splits the 3D space into tokens, with each token's predicted volume contributing to the final 3D reconstruction. Three model variations are also implemented: VolT, VolT+, and EVolT. EVolT's design incorporates a novel approach that integrates a view-divergence enhancement function aimed at mitigating divergence decay within its self-attention layers. Despite very small trainable parameters, this model achieved state-of-the-art results, especially with multi-view samples. 


\par 
Despite these advancements, several research gaps persist. While many current methods prioritize enhancing feature fusion and attention mechanisms, they often do not fully tackle challenges such as handling noisy or sparse input data or mitigating the impact of input order dependency on model accuracy. To address these issues, our approach utilizes an attention network to effectively aggregate features from input images. Our hybrid network combines efficient CNN feature representations with precise self-attention, aiming to improve the resilience and accuracy of 3D model reconstruction. The specifics of the proposed framework are outlined in the upcoming section.

\section{Methodology}
Our method introduces a technique to generate 3D models from RGB images, both single and multi-view, utilizing a predicted grid of voxel occupancy in three dimensions. The architecture of the network is depicted in Fig.~\ref{fig_fig3}. Our method introduces a novel network architecture for 3D model reconstruction, complemented by a dedicated training algorithm, JTSO, designed to improve efficiency. The architecture begins with a shared CNN encoder-decoder framework, followed by a 3D U-Net for refining features. Addressing feature fusion challenges, we incorporate an attention network inspired by Transformer models to effectively transform latent space features. For robust feature extraction and object generation, we employ a ResNet18-based autoencoder architecture. Additionally, refiner networks based on 3D U-Net optimize and enhance the 3D structures generated by decoders. Mean cross-entropy loss from intermediate and refined 3D models is used to update the encoder-decoder and attention module. We used the Adam optimizer with a step-wise decreasing learning rate. Our proposed network is trained with a dedicated algorithm called JTSO. Detailed descriptions of each module used in our architecture are provided in the following sections.

\subsection{\textbf{Encoder}}
The input images undergo processing through an encoder network designed to extract a specific set of deep image features using 2D convolution layers. We utilize the ResNet-V1 architecture \cite{b16}, comprising 12 layers, to extract 1024 feature vectors from images sized $127 \times 127 \times 3$. Similar encoder-decoder configurations are employed in state-of-the-art methods such as 3D-R2N2 \cite{b2} and AttSets \cite{b1}. The encoder network consists of 12 convolution layers organized into 6 residual blocks. Each residual block includes 2 convolutional blocks with the Leaky ReLU activation function, followed by a max-pool layer and an identity connection. Identity connections incorporate a 2D convolution layer with a 1x1 kernel size. The output channel sizes of the residual blocks are 96, 128, 256, 256, 256, and 256 respectively. All max-pool layers have a kernel size and stride of 2, and identity connections follow each residual block except for the 4th block. Ultimately, the encoder network produces flattened feature vectors consisting of 1024 dimensions.


\subsection{\textbf{Self attention module}}


\par
The transformer-inspired self-attention network \cite{b19} aims to enrich extracted features by enabling the model to selectively emphasize various segments of the input sequence, thereby comprehending their interconnectedness. The self-attention module utilizes a multi-head attention mechanism, calculating attention weights and producing a contextual vector for each element in the input. This module uses eight attention heads and processes a latent vector of size 1024, ensuring a comprehensive capture of feature interactions. Each attention head calculates scaled dot product attention as given below.


Using the matrices for query ($\mathbf{Q}$), key ($\mathbf{K}$), and value ($\mathbf{V}$), the scaled dot-product attention is calculated in the following manner:

\begin{equation}
\text{Attention}(\mathbf{Q}, \mathbf{K}, \mathbf{V}) = \text{softmax}\left(\frac{\mathbf{Q}\mathbf{K}^\top}{\sqrt{d_k}}\right)\mathbf{V}
\end{equation}

Here, $d_k$ represents the dimensionality of the key vectors.

\par
The self-attention mechanism first projects the input latent vector into query, key, and value matrices. Each attention head independently computes scaled dot-product attention, producing attention scores that are subsequently used to weigh the input features. The outputs from all attention heads are then concatenated and linearly transformed to produce the final attended feature representation using the weight matrix $\mathbf{W}^O$.  This approach allows the network to jointly attend to information from different representation subspaces, enhancing the overall feature representation. This can be represented as follows,

\begin{figure}[h]
\centering
\includegraphics[width=\linewidth]{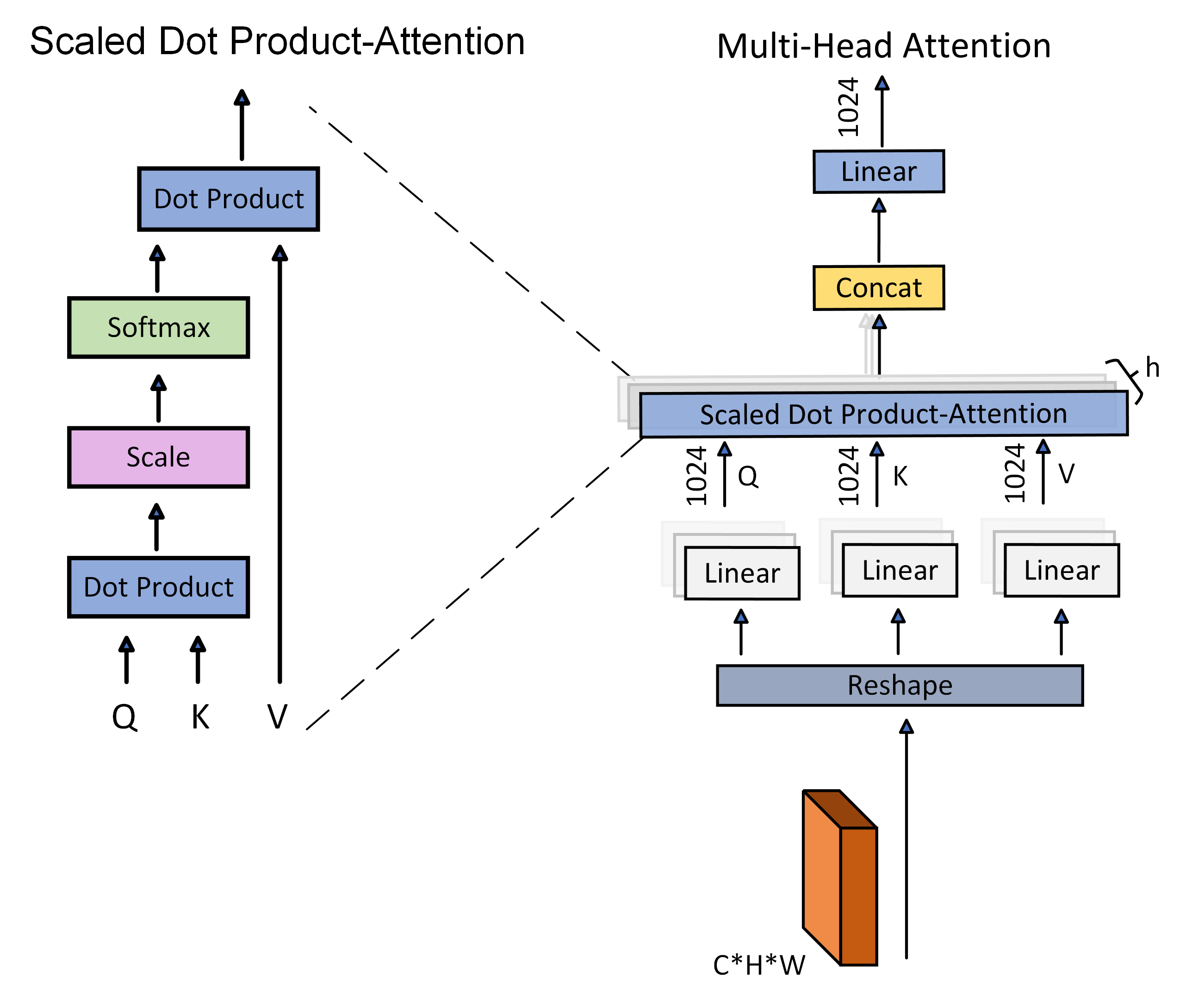}
\caption{Illustration of multi-head attention and Scaled dot-product attention}
\label{fig_fig4}
\end{figure}

\begin{equation}
\text{MultiHead}(\mathbf{Q}, \mathbf{K}, \mathbf{V}) = \text{Concat}(\text{head}_1, \text{head}_2, \ldots, \text{head}_h) \mathbf{W}^O,
\end{equation}
\begin{equation}
\text{where} \quad \text{head}_i = \text{Attention}(\mathbf{Q}\mathbf{W}_i^Q, \mathbf{K}\mathbf{W}_i^K, \mathbf{V}\mathbf{W}_i^V).
\end{equation}

Here:
\begin{align*}
\mathbf{W}_i^Q &\in \mathbb{R}^{d_{model} \times d_q}, \\
\mathbf{W}_i^K &\in \mathbb{R}^{d_{model} \times d_k}, \\
\mathbf{W}_i^V &\in \mathbb{R}^{d_{model} \times d_v}, \\
\mathbf{W}^O &\in \mathbb{R}^{hd_v \times d_{model}}.
\end{align*}
Were, $\mathbf{W}_i^Q$, $\mathbf{W}_i^K$, and $\mathbf{W}_i^V$ are the weight matrices for the queries, keys, and values for head $i$, respectively. These matrices project the input dimensions ($d_{model}$) into smaller dimensions ($d_q$, $d_k$ and $d_v$). $\mathbf{W}^O$ is the weight matrix used to combine the concatenated outputs of all heads back into the model's dimension ($d_{model}$).

\par
The input and output latent vectors of the self-attention module maintain the same size, preserving the dimensional consistency required for subsequent processing stages. By incorporating this self-attention mechanism into the auto-encoder, which accepts multiple images as input and creates a latent vector of size 1024, we achieve a more robust and powerful feature representation in the latent space. Furthermore, the self-attention mechanism's capacity to simultaneously attend to multiple aspects of the input features leads to a richer and more comprehensive understanding of complex shapes. This significantly improves low-level image feature processing methods that struggle with intricate geometries.

\subsection{\textbf{Decoder}}
The decoder generates 3D models from the aggregated feature output produced by the Attention-Net block. It takes as input a 1024-dimensional vector obtained from the self-attention layer. The decoder consists of five residual blocks with output filter sizes of 128, 128, 128, 64, 64, and 32, respectively. Each transpose convolution layer in the decoder has a kernel size of 3, which progressively upsamples the feature maps. Additionally, each identity connection within the residual blocks incorporates a 2D convolution layer with a kernel size of 1 to maintain consistent dimensions and facilitate efficient gradient flow. This design enables the decoder to effectively reconstruct detailed 3D shapes from the high-dimensional feature vector.

\subsection{\textbf{Refiner Module}}
The architecture of the Refiner module is adapted from a 3D U-Net architecture. This module aids in optimizing the predicted 3D volumes. Following an encoder-decoder architecture, the Refiner module differs from the conventional setup by incorporating a U-Net connection instead of a residual connection. This U-Net connection ensures the preservation of the predicted 3D volume from the decoder. The input and output dimensions of the Refiner are 32 $^3$ voxels.

The Refiner's encoder consists of three sequential 3D convolutional layers. Each convolutional layer utilizes a filter size of $4 \times 4 \times 4$ and a padding of 2. Following each convolution, a batch normalization layer is applied, along with a leaky ReLU activation function. To reduce dimensionality, a max pooling layer with a kernel size of $2 \times 2 \times 2$ is employed after each convolution. The number of output channels progressively increases through the encoder, starting with 32 in the first layer, 64 in the second, and finally reaching 128 in the third layer.

The decoder, in contrast, leverages three transposed convolutional layers. These layers each possess a filter bank of $4 \times 4 \times 4$ filters and a padding of 2, but with a stride of 1 to maintain spatial resolution. Batch normalization and ReLU activation are incorporated after each transposed convolution. The final layer in the decoder utilizes a sigmoid function to generate the output.

\begin{figure}[h]
\centering
\includegraphics[width=\linewidth]{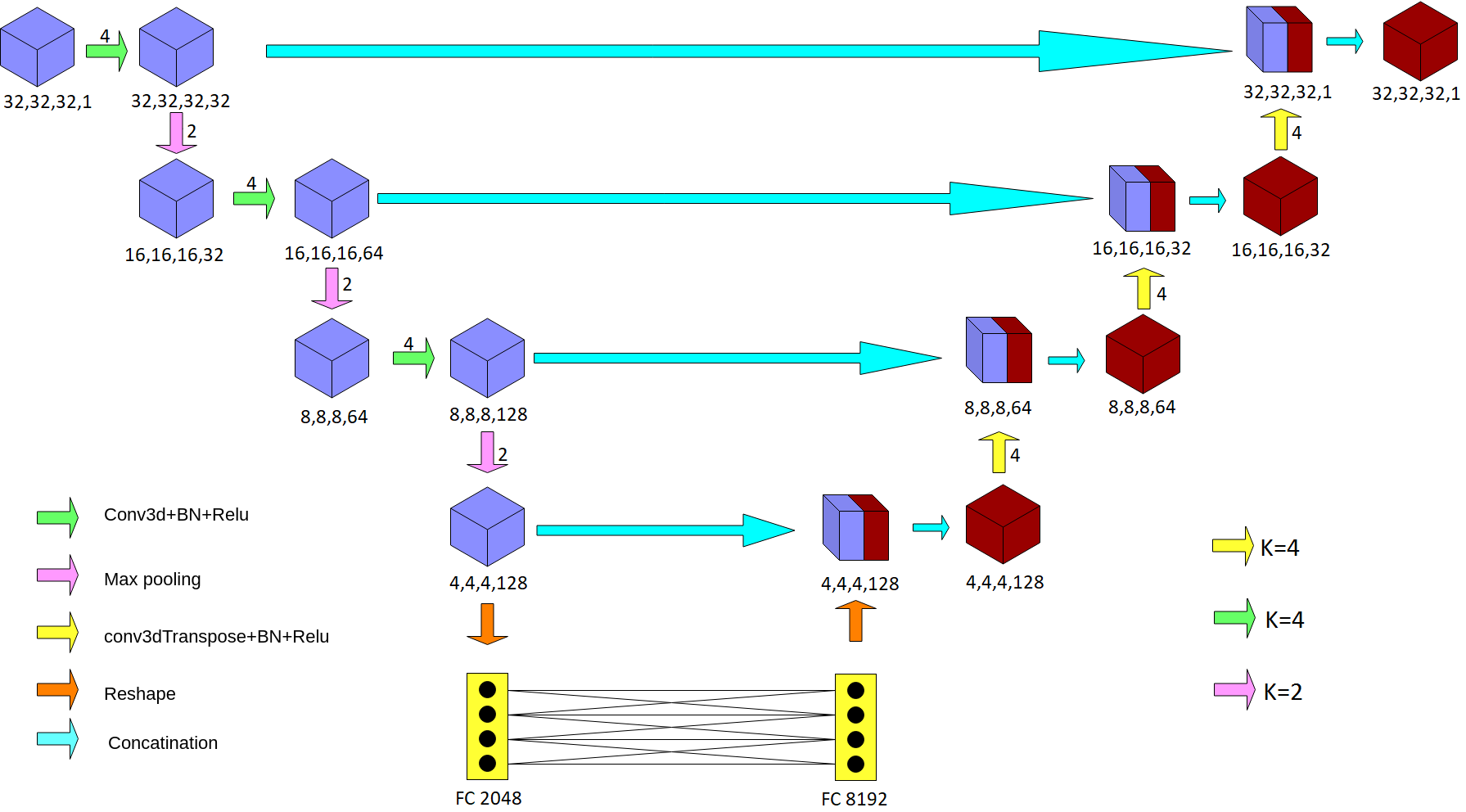}
\caption{Detailed architecture of Refiner Network}
\label{fig_fig6}
\end{figure}


\section{\textbf{Training Algorithm}}
 

\par
The traditional end-to-end training methods do not ensure that the encoder-decoder network and self-attention module \cite{b1} can learn basic visual features and corresponding self-attention scores separately. Due to varying network-generated losses and the absence of distinct feature score labels for separate training, the feasibility of end-to-end joint training hinges on consistent input image counts. Should both encoder-decoder and attention modules undergo joint training with multiple input images, the encoder's focus naturally shifts towards extracting features across these images. Consequently, the network output will degrade if the number of input images is reduced to one. Furthermore, the refiner network layers also update with incorrect weight variables. This is because the refiner network does not depend on the number of input images, and unlike others, its input and output are 3D.

\begin{figure}[h]
\centering
\includegraphics[width=\linewidth]{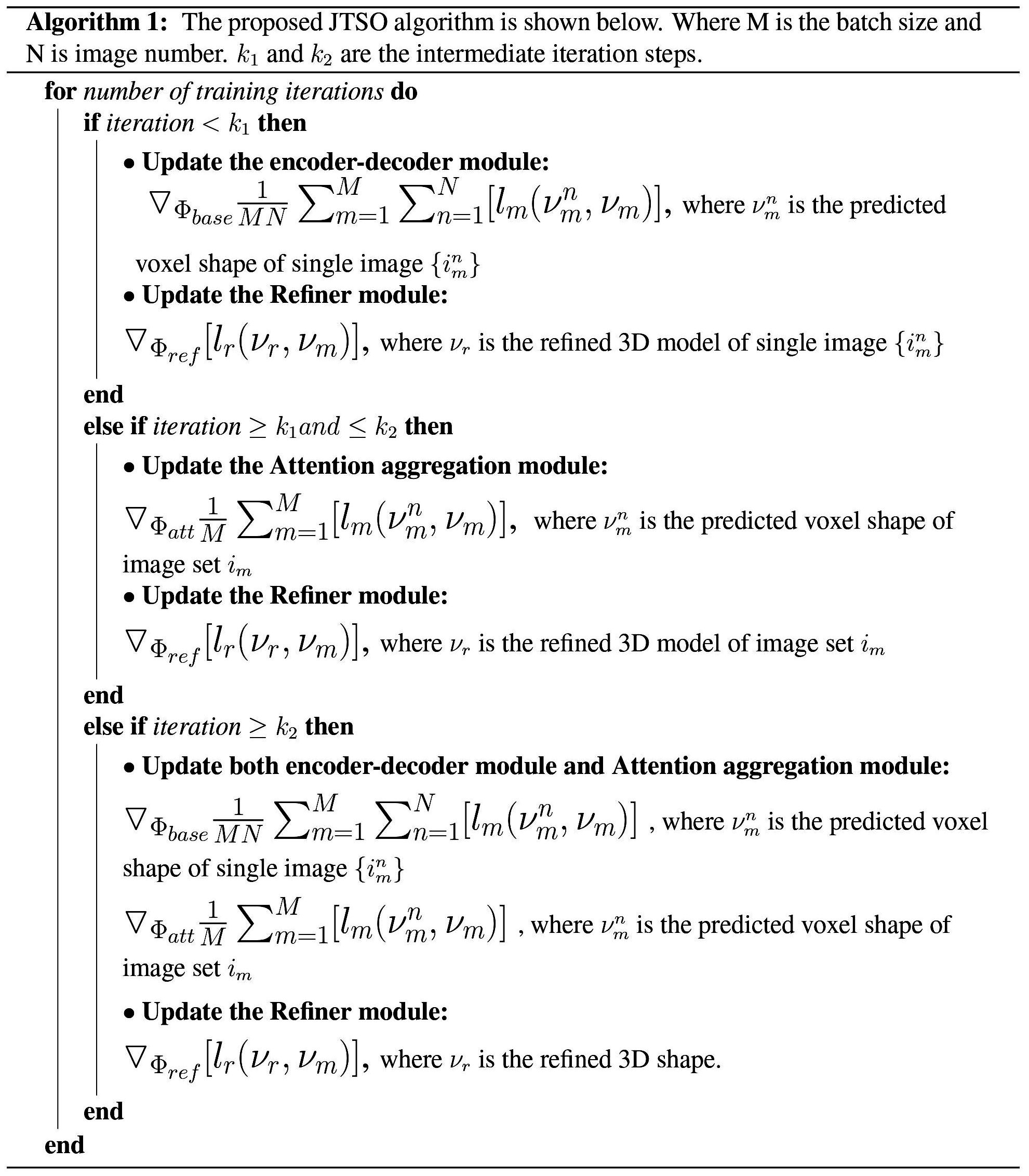}
\caption{Proposed JTSO Algorithm flow}
\label{fig_fig5}
\end{figure}


\par 
To address these issues, we propose a \textbf{JTSO (Joint Train Separate Optimize)} algorithm, inspired by concepts from \citeauthor{b1}\cite{b1}, which addresses separate optimization needs in networks. The JTSO algorithm optimizes encoder-decoder network layers when there is one input image and optimizes the self-attention module when multiple input images are present. This decoupling strategy enhances the robustness of deep feature learning in encoder-decoder networks and improves the generalization of attention scores in the self-attention module across feature sets.

\par
The proposed network generates two 3D models for each input: one from the encoder-decoder and another from the refiner network. Both models are updated using mean cross-entropy loss compared to ground truth. The training algorithm has three phases. In the first phase, the encoder-decoder is updated until convergence. In the second phase, the self-attention module is updated with cross-entropy loss, and the refiner module is updated with the refiner output loss in both phases. The final phase, called Joint Train Separate Optimize (JTSO), updates both the encoder-decoder and self-attention modules using single and multiple images of the same class, respectively.

\par
The encoder-decoder network's trainable parameters are labeled as $\Theta_{base}$, while the self-attention module's trainable parameters are represented by $\Phi_{att}$, and those of the refiner network by $\Phi_{ref}$. The mean loss ($l_{m}$)  is calculated from the predicted voxel loss ($l_{p}$) and the refined voxel loss ($l_{r}$). It can be seen that $\Theta_{base}$ and $\Phi _{att}$ are separately optimized in two stages. Our proposed JTSO algorithm is outlined in figure \ref{fig_fig5}. 
\par


\subsection{\textbf{Loss Function}}






The network employs a loss function that calculates the average voxel-wise cross-entropy between the reconstructed model and the corresponding ground truth, represented succinctly as follows:

\begin{equation}
\mathcal{L} = \frac{1}{V} \sum_{v=1}^{V} \left[ \mathrm{GT}_{(v)} \log(\hat{p}_{(v)}) + (1 - \mathrm{GT}_{(v)}) \log(1 - \hat{p}_{(v)}) \right]
\end{equation}

where \( V \) denotes the total number of voxels in the ground truth data, \( \hat{p}_{(v)} \) signifies the predicted occupancy probability for voxel \( v \), and \( \mathrm{GT}_{(v)} \) represents the corresponding ground truth value for the voxel. A lower value of \( \mathcal{L} \) signifies a prediction that is more accurately aligned with the ground truth.






\subsection{\textbf{Dataset and Metrics}}

\textbf{ShapeNet:} The ShapeNet \cite{b6} dataset aggregates a wide array of 3D CAD models, meticulously categorized based on the WordNet hierarchy. For our experiments, we utilized a subset of the ShapeNet dataset \cite{b20}, which comprises 16,896 models spanning five major categories. The dataset was divided into training and testing sets with a split ratio of 80:20, ensuring a robust evaluation of our model. Each 3D model comes with images that have been taken from 24 different viewpoints. Refer Appendix-A for qualitative images. 

To assess the performance of the 3D reconstruction, we employed the \textbf{Intersection over Union (IoU)} metric \cite{b32}. This metric is a standard for evaluating the accuracy of object reconstruction by comparing the predicted voxel occupancy with the ground truth voxel occupancy. Specifically, we binarized the predicted probabilities using a fixed threshold of 0.25 to determine voxel occupancy. The IoU is calculated as follows:

\begin{equation}
IOU = \frac{\sum_{i,j,k} I_{(p_{(i,j,k)} > t)} I_{(gt_{(i,j,k)})}}{\sum_{i,j,k} I[I_{(p_{(i,j,k)} > t)} I_{(gt_{(i,j,k)})}]}
\end{equation}

In this equation, \( p_{(i,j,k)} \) represents the predicted probability of occupancy at the voxel coordinate \((i, j, k)\), and \( gt_{(i,j,k)} \) denotes the ground truth occupancy at the same coordinate. The function \( I(\cdot) \) is an indicator function that returns 1 if the condition inside is true and 0 otherwise, and \( t \) is the voxelization threshold. Higher IoU values signify more precise reconstruction outcomes, demonstrating how closely the predicted voxel occupancies align with the ground truth.

The choice of IoU as our evaluation metric allows us to quantitatively measure the fidelity of our reconstructed 3D models against the actual models, providing a clear and objective means of assessing the effectiveness of our approach.

\subsection{\textbf{System configuration details}}
We utilize $127 \times 127$ RGB images as input to train the proposed network, with a batch size of 4. The voxelized output model is produced with a resolution of $32^3$. Our implementation leverages TensorFlow 2.8.0 and Python 3.9.8, alongside CUDA 12.2 and cuDNN 8.4 as the backend driver and library, respectively. Training is performed on a system equipped with $4 \times RTX 3090 $ GPUs, with the entire training process taking 2 to 3 days, depending on specific settings.

Mixed-precision computation is employed to accelerate the training process. The optimization is conducted using the Adam optimizer, with $\beta_1$ set to 0.9 and $\beta_2$ set to 0.999. The learning rate starts from 0.001 and undergoes a decay factor of 2 after 150 epochs. 

\section{Results}
In this section, we present the quantitative results of various methods on the ShapeNet dataset using Intersection over Union (IoU) evaluation metrics. The methods evaluated include those based on CNNs, RNNs, and transformers. To ensure a fair comparison, all methods were tested using the same set of input images across all experiments.

\subsection{\textbf{Reconstruction of Synthetic images}}
In Table \ref{tab:label1}, you can find the Mean IOU scores per category from single-view reconstruction using the ShapeNet testing dataset. Our model exhibits notably better performance across various categories when compared to current leading methods.

The proposed model shows notable enhancements across various categories: Airplane (+1.0\%), Bench (+4.5\%), Cabinet (+2.1\%), Car (+1.0\%), Chair (+8.9\%), Display (+5.3\%), Lamp (+15.7\%), Speaker (+2.7\%), Rifle (+6.1\%), Sofa (+4.6\%), Table (+7.3\%), Telephone (+5.2\%), and Watercraft (+1.5\%).
Notably, the proposed model achieves the highest mean IOU overall, outperforming other models by 4.2\%. These results suggest that the proposed model offers substantial improvements in single-view reconstruction accuracy compared to existing state-of-the-art methods, making it a promising approach for various object categories.

\begin{table*}[h]
\centering
\caption{Average IOU per category in reconstructing single views using the ShapeNet test dataset.}
\label{tab:label1}
\begin{tabularx}{\linewidth}{lXXXXXXXXX}

\toprule
Category    & 3D-R2N2 \cite{b2} & Attsets \cite{b1} & Pix2Vox-F \cite{b4} & Pix2Vox-A  \cite{b4} &  VolT  \cite{b31}     &  EVolT  \cite{b31}    & Proposed \\ 
\midrule
Airplane    & 0.513   & 0.594   & 0.600     & 0.684     & 0.562     & 0.562     & \textbf{0.691} \\ 
Bench       & 0.421   & 0.552   & 0.538     & 0.616     & 0.511     & 0.511     & \textbf{0.644} \\ 
Cabinet     & 0.716   & 0.783   & 0.765     & 0.792     & 0.713     & 0.713     & \textbf{0.809} \\
Car         & 0.798   & 0.844   & 0.837     & 0.854     & 0.785     & 0.785     & \textbf{0.863} \\
Chair       & 0.466   & 0.559   & 0.535     & 0.567     & 0.541     & 0.541     & \textbf{0.618} \\
Display     & 0.468   & 0.565   & 0.511     & 0.537     & 0.576     & 0.576     & \textbf{0.607} \\
Lamp        & 0.381   & 0.445   & 0.435     & 0.443     & 0.456     & 0.456     & \textbf{0.528} \\
Speaker     & 0.662   & 0.721   & 0.707     & 0.714     & 0.664     & 0.664     & \textbf{0.741} \\
Rifle       & 0.544   & 0.601   & 0.598     & 0.615     & 0.610     & 0.630     & \textbf{0.653} \\
Sofa        & 0.628   & 0.703   & 0.687     & 0.709     & 0.649     & 0.659     & \textbf{0.742} \\
Table       & 0.513   & 0.590   & 0.587     & 0.601     & 0.586     & 0.586     & \textbf{0.645} \\
Telephone   & 0.661   & 0.743   & 0.770     & 0.776     & 0.692     & 0.692     & \textbf{0.817} \\
Watercraft  & 0.513   & 0.601   & 0.582     & 0.594     & 0.577     & 0.577     & \textbf{0.603} \\

\bottomrule
Overall     & 0.560   & 0.642   & 0.634     & 0.661     & 0.605     & 0.609     & \textbf{0.689} \\

\bottomrule
\end{tabularx}
\end{table*}

\begin{table}
\centering
\caption{Per-Category Mean IOU for Multi-View Reconstruction on ShapeNet Testing Dataset}
\label{tab:label2}
\begin{tabularx}{\linewidth}{lXXXXXXXXXXX}
\toprule
Methods & 1 view & 3 view & 5 view & 8 view & 12 view & 15 view & 18 view & 20 view\\ 
\midrule
3D-R2N2 & 0.560 & 0.617 & 0.625 & 0.634 & 0.635 & 0.636 & 0.636 & 0.636 \\ 
Attsets & 0.642 & 0.670 & 0.677 & 0.685 & 0.688 & 0.692 & 0.693 & 0.693  \\
Pix2Vox-F & 0.634 & 0.660 & 0.668 & 0.673 & 0.676 & 0.680 & 0.682 & 0.684  \\
Pix2Vox-A & 0.661 & 0.686 & 0.693 & 0.697 & 0.699 & 0.702 & 0.704 & 0.704 \\

VolT   & 0.605 & 0.624 & 0.662 & 0.681 & 0.699 & 0.706 & 0.708 & 0.711  \\
EVolT & 0.609 & 0.637 & 0.671 & 0.698 & 0.720 & \textbf{0.729} & \textbf{0.732} & \textbf{0.735}  \\

Proposed  & \textbf{0.689} & \textbf{0.698} & \textbf{0.706} & \textbf{0.713} & \textbf{0.721} & 0.727 & 0.730 & 0.733  \\

\bottomrule
\end{tabularx}
\end{table}
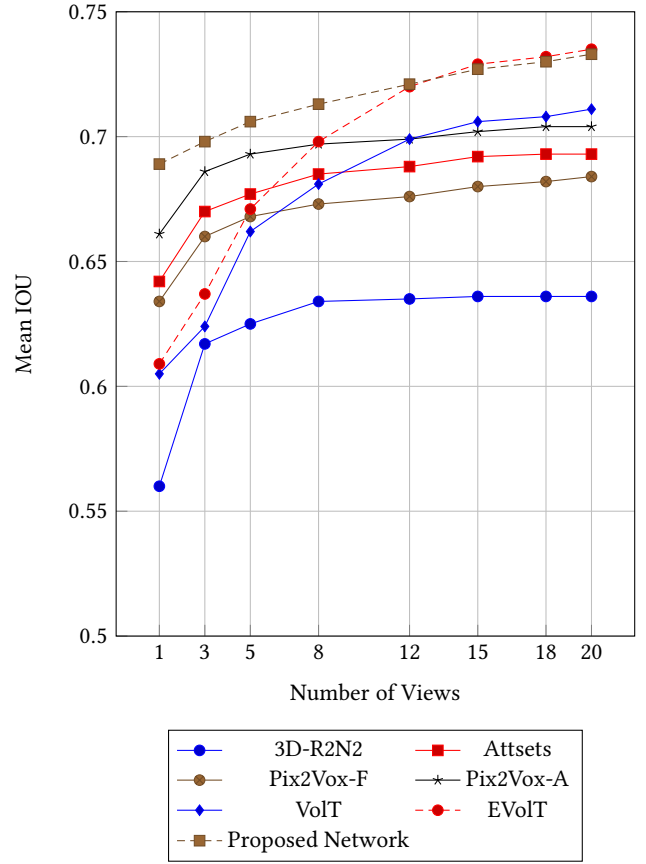
\begin{figure}[!htb]
\centering
\begin{tikzpicture}
\begin{axis}[
    width=\linewidth,
    height=\linewidth+\linewidth/6,
    xlabel={Number of Views},
    ylabel={Mean IOU},
    xtick={1,3,5,8,12,15,18,20},
    legend style={at={(0.5,-0.15)},anchor=north,legend columns=2},
    grid=major,
    ymin=0.5, ymax=0.75,
    xtick pos=left,
    ytick pos=left,
    mark options={solid}
]
\addplot coordinates {(1,0.560) (3,0.617) (5,0.625) (8,0.634) (12,0.635) (15,0.636) (18,0.636) (20,0.636)};
\addplot coordinates {(1,0.642) (3,0.670) (5,0.677) (8,0.685) (12,0.688) (15,0.692) (18,0.693) (20,0.693)};
\addplot coordinates {(1,0.634) (3,0.660) (5,0.668) (8,0.673) (12,0.676) (15,0.680) (18,0.682) (20,0.684)};
\addplot coordinates {(1,0.661) (3,0.686) (5,0.693) (8,0.697) (12,0.699) (15,0.702) (18,0.704) (20,0.704)};
\addplot coordinates {(1,0.605) (3,0.624) (5,0.662) (8,0.681) (12,0.699) (15,0.706) (18,0.708) (20,0.711)};
\addplot coordinates {(1,0.609) (3,0.637) (5,0.671) (8,0.698) (12,0.720) (15,0.729) (18,0.732) (20,0.735)};
\addplot coordinates {(1,0.689) (3,0.698) (5,0.706) (8,0.713) (12,0.721) (15,0.727) (18,0.730) (20,0.733)};
\legend{3D-R2N2, Attsets, Pix2Vox-F, Pix2Vox-A, VolT, EVolT, Proposed Network}
\end{axis}
\end{tikzpicture}
\caption{Average Intersection over Union (IOU) by Category in ShapeNet's Test Set for Multi-View 3D Reconstruction.}
\label{fig:linechart}
\end{figure}


Table \ref{tab:label2} shows mean IOU per-category scores of different approaches of multi-view reconstruction on the ShapeNet test dataset 
The proposed network performs superiorly to other models in the multi-view 3D reconstruction task on the ShapeNet test dataset, as evidenced by the per-category mean intersection over union (IOU) scores across various views. Starting with a single view, the Proposed Network achieves a Mean IOU of 0.689, outperforming all other models, with the next best, Pix2Vox-A, at 0.661. This trend continues across increasing numbers of views, maintaining the highest Mean IOU scores at each interval. By the 3-view mark, the Proposed Network reaches 0.698, again leading over the closest competitor, Pix2Vox-A, which scores 0.686. As the number of views increases to 8, the Proposed Network secures 0.713, surpassing EVolT's 0.698. Refer Appendix-B \& C for qualitative results. 

\par

The Proposed Network demonstrates strong performance up to 20 views with a Mean IOU of 0.733, slightly below EVolT's 0.735. This achievement leverages an attention network for effective feature aggregation from input images. Additionally, a two-stage 3D model generation process enhances the precision of 3D model representations. The hybrid network integrates CNN feature representations with precise transformer feature attention aggregation, ensuring consistent high performance across all views. This robustness and effectiveness in multi-view reconstruction tasks signify a significant advancement over existing methods like 3D-R2N2 \cite{b2}, Attsets \cite{b1}, Pix2Vox \cite{b4}, VolT, and EVolT \cite{b31}, particularly in initial and mid-range view scenarios.

\section{\textbf{Effect of Architectural Enhancements on Performance}}
\subsection{\textbf{Impact of Refiner Network}}
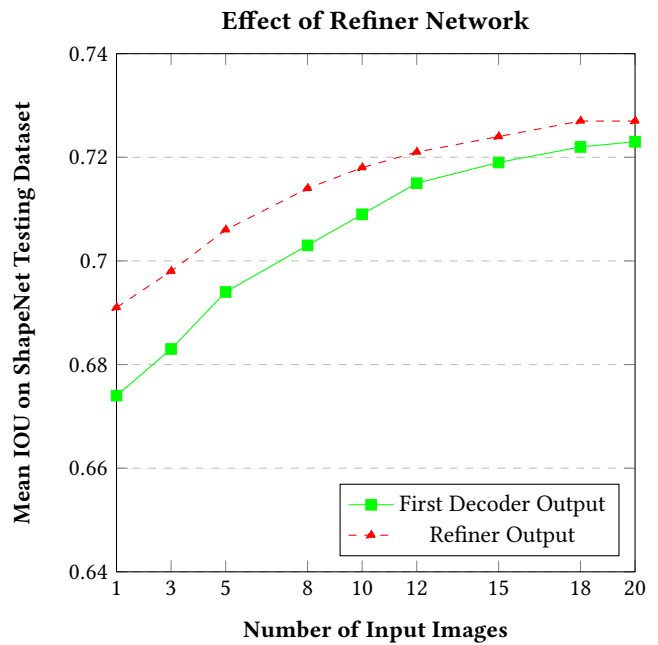
\begin{figure}[h]
    \centering
    \label{graph2}
    \begin{tikzpicture}[thick,scale=1.0]
        \begin{axis}[
            title={\textbf{\large{Effect of Refiner Network}}},
            xlabel={\textbf{Number of Input Images}},
            ylabel={\textbf{Mean IOU on ShapeNet Testing Dataset}},
            xmin=1, xmax=20,
            ymin=.64, ymax=0.74, 
            xtick={1,3,5,8,10,12,15,18,20},
            ytick={0.64,0.66,0.68,0.70,0.72,0.74}, 
            legend pos=south east, 
            ymajorgrids=true,
            grid style=dashed,
            width=\linewidth,
            height=\linewidth,
            cycle list={%
                {green, solid, mark=square*}, 
                {red, dashed, mark=triangle*} 
            }
        ]
            \addplot
                coordinates {
                (1,0.674)(3,0.683)(5,0.694)(8,0.703)(10,0.709)(12,0.715)(15,0.719)(18,0.722)(20,0.723)
                };
            \addplot
                coordinates {
                (1,0.691)(3,0.698)(5,0.706)(8,0.714)(10,0.718)(12,0.721)(15,0.724)(18,0.727)(20,0.727)
                };
            \legend{First Decoder Output, Refiner Output}
        \end{axis}
    \end{tikzpicture}
    \caption{The evaluation of IoU is influenced by both the refiner network's effectiveness and the variety of views used in the study.}
    \label{fig:graph-5}
\end{figure}

In this section, we evaluate the effect of the refiner network by comparing the output of the refiner and auto-encoder using the Mean IOU of the ShapeNet testing dataset. However, from figure \ref{fig:graph-5}, it is clear that adding more views weakens the impact of the refiner network.  

\subsection{\textbf{Impact in Resource Utilization}}
The table \ref{tab:memory_usage} compares the memory utilization and inference time of the proposed network with several state-of-the-art architectures on the ShapeNet dataset, tested on an RTX 8000 48GB GPU. The proposed network has 143 million parameters, which is substantially higher than the other models listed, indicating a more complex architecture. Its memory usage stands at 2900 MB, again the highest among the compared methods, reflecting its extensive computational requirements. However, the proposed network shows efficiency in training time, requiring only 24 hours, which is competitive given its complexity. For inference, the backward pass time is 78.12 ms, suggesting a balanced performance in terms of computational speed and resource utilization. The proposed network demonstrates robust performance with optimized memory utilization, despite its larger parameter count and memory usage.

\begin{table}
\centering
\caption{Memory footprint and inference time on ShapeNet dataset.}
\label{tab:memory_usage}
\begin{tabularx}{\linewidth}{l*{5}{X}}
\toprule
Methods & 3D-R2N2  & Pix2Vox-F & Pix2Vox-A & EVolT & Proposed\\ 
\midrule
\#Parameters (M) & 35.97  & 7.41 & 114.24 & 29.03 & 143\\ 
Memory (MB) & 1407  & 673 & 2729 & 980 & 2900\\ 
\midrule
Training (hours) & 169  & 12 & 25 & 10 & 24\\ 
Forward (ms) & 312.50  & 12.93 & 72.01 & 22 & 78.12\\ 
\bottomrule
\end{tabularx}
\end{table}
\section{Conclusion}
In this paper, we propose a unique approach for 3D model reconstruction from single and multi-view images. The proposed network demonstrates robust performance in multi-view 3D reconstruction tasks, particularly in scenarios involving fewer input images. The architecture outperformed other SOTA models by 4.2\% in single-view 3D reconstruction. This approach shows consistent results with arbitrary input order.

In upcoming research, our emphasis will shift towards refining the precision and detail of our 3D models. This will involve enhancing the feature vectors employed within the attention layer. A more sophisticated selection of feature vectors could significantly improve the network's ability to handle multiple input images, thereby boosting overall accuracy. Additionally, various improvements in the self-attention mechanism, particularly with multi-dimensional vectors, could further optimize performance. This adjustment would likely enhance the model's capability to process and integrate features from numerous input images effectively.










\end{document}